\documentclass[10pt,twocolumn,letterpaper]{article}
\usepackage[table]{xcolor}  

\usepackage{cvpr}              %

\usepackage[utf8]{inputenc} %
\usepackage[T1]{fontenc}    %
\usepackage{url}            %
\usepackage{booktabs}       %
\usepackage{amsfonts}       %
\usepackage{nicefrac}       %
\usepackage{microtype}      %
\usepackage{graphicx}
\usepackage{wrapfig}
\makeatletter
\@namedef{ver@everyshi.sty}{}
\makeatother
\usepackage{tikz}
\usepackage{enumitem}%

\usepackage{amsmath}

\usepackage{url}

\usepackage[pagebackref,breaklinks,colorlinks,bookmarks=false]{hyperref}

\usepackage{booktabs}
\usepackage{multicol}
\usepackage{multirow}
\usepackage{color}
\usepackage{caption}
\usepackage{hhline}
\usepackage{pifont}
\usepackage{threeparttable}
\usepackage{makecell}
\usepackage{algorithm} 
\usepackage{listings}
\usepackage{amsfonts,amssymb}
\usepackage{graphicx}
\usepackage{pifont}%
\newcommand{\cmark}{\textcolor{green}{\ding{51}}}%
\newcommand{\xmark}{\textcolor{red}{\ding{55}}}%

\definecolor{attcolor}{rgb}{0. , 0.5 , 0.9}
\definecolor{convcolor}{rgb}{1 , 0.7 , 0}
\definecolor{mlpcolor}{rgb}{0.9 , 0.3 , 0.4}
\definecolor{othercolor}{rgb}{0.8 , 0.4 , 0.9}
\definecolor{attconvcolor}{rgb}{0. , 0.6 , 0.1}
\definecolor{wihtecolor}{rgb}{1 , 1. , 1.}
\definecolor{blackcolor}{rgb}{0, 0. , 0.}
\definecolor{RowColor}{rgb}{0.97, 0.97, 1}

\newcommand{\attshape}{\raisebox{0.5pt}{\tikz\fill[attcolor] (0,0) circle (.8ex);}}
\newcommand{\convshape}{\raisebox{0.5pt}{\tikz\fill[convcolor] (0,0) circle (.8ex);}}
\newcommand{\mlpshape}{\raisebox{0.5pt}{\tikz\fill[mlpcolor] (0,0) circle (.8ex);}}
\newcommand{\convattshape}{\raisebox{0.5pt}{\tikz\fill[attconvcolor](0,0) circle (.8ex);}}
\newcommand{\othershape}{\raisebox{0.5pt}{\tikz\fill[othercolor] (0,0) circle (.8ex);}}
\newcommand{\whiteshape}{\raisebox{0.5pt}{\tikz\fill[wihtecolor] (0,0) circle (.8ex);}}

\def\ie{\emph{i.e.}}
\def\eg{\emph{e.g.}}
\def\etc{\emph{etc}}
\def\etal{{\em et al.~}}

\usepackage[capitalize]{cleveref}
\crefname{section}{Sec.}{Secs.}
\Crefname{section}{Section}{Sections}
\Crefname{table}{Table}{Tables}
\crefname{table}{Tab.}{Tabs.}

\def\cvprPaperID{4540} %
\def\confName{CVPR}
\def\confYear{2023}

\begin{document}

\title{A Close Look at Spatial Modeling: From Attention to Convolution}

\author{
Xu Ma\textsuperscript{$\dagger$}, 
Huan Wang\textsuperscript{$\dagger$}, 
Can Qin\textsuperscript{$\dagger$}, 
Kunpeng Li\textsuperscript{$\ddagger$}, 
Xingchen Zhao\textsuperscript{$\dagger$}, 
Jie Fu\textsuperscript{$\sharp$}, 
Yun Fu\textsuperscript{$\dagger$}
\\
\textbf{
\textsuperscript{$\dagger$}Northeastern University \qquad
\textsuperscript{$\ddagger$}Meta Reality Labs \qquad
\textsuperscript{$\sharp$}Mila\qquad
}
}
\maketitle

\begin{abstract}
Vision Transformers have shown great promise recently for many vision tasks due to the insightful architecture design and attention mechanism.
By revisiting the self-attention responses in Transformers, we empirically observe two interesting issues. First, Vision Transformers present a query-irrelevant behavior at deep layers, where the attention maps exhibit nearly consistent contexts in global scope, regardless of the query patch position (also head-irrelevant).
Second, the attention maps are intrinsically sparse, few tokens dominate the attention weights; introducing the knowledge from ConvNets would largely smooth the attention and enhance the performance.
Motivated by above observations, we generalize self-attention formulation to abstract a query-irrelevant global context directly and further integrate the global context into convolutions.
The resulting model, a Fully Convolutional Vision Transformer (\textit{i.e.}, FCViT), purely consists of convolutional layers and firmly inherits the merits of both attention mechanism and convolutions, including dynamic property, weight sharing, and short- and long-range feature modeling, \etc{}. 
Experimental results demonstrate the effectiveness of FCViT. With less than 14M parameters, our FCViT-S12 outperforms related work ResT-Lite by 3.7\% top-1 accuracy on ImageNet-1K. When scaling FCViT to larger models, we still perform better than previous state-of-the-art ConvNeXt with even fewer parameters.  
FCViT-based models also demonstrate promising transferability to downstream tasks, like object detection, instance segmentation, and semantic segmentation.
Codes and models are made available at: \href{https://github.com/ma-xu/FCViT}{https://github.com/ma-xu/FCViT}.
  
\end{abstract}

\section{Introduction}
\label{sec:introduction}
\begin{figure}
    \centering
    \includegraphics[width=1\linewidth]{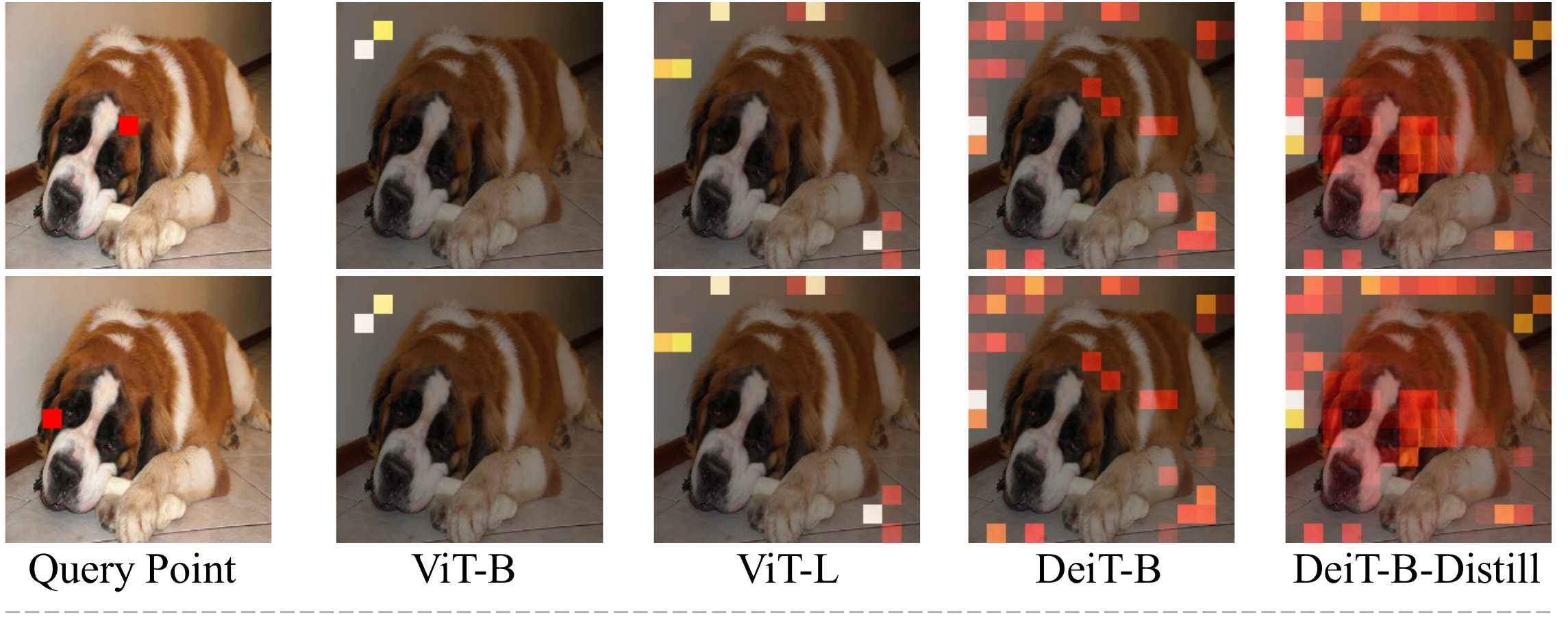}
    \hspace{2mm}
    \includegraphics[width=1\linewidth]{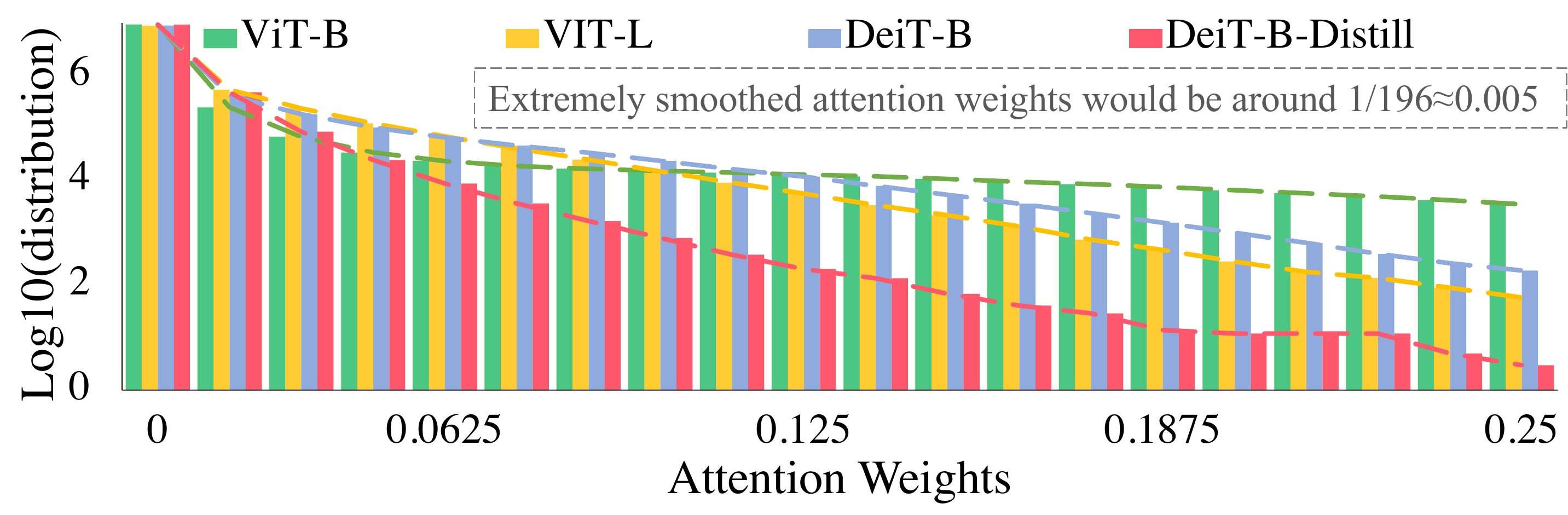}
    \caption{
    Illustration of sparse and query-irrelevant issues in deep attention layers. 
    We examine several ViT variants (patch size = 16 and image patches number =196 for all).
    \textbf{Top:} examples of two different query points and related attention maps.
    \textbf{Bottom:} statistical analysis of the distribution of attention weights over ImageNet-1K validation set, normalized by Log10. If most patches contribute to attention, the attention weights would be concentrated at a very small value ($\sim$0.005) and marginally large values; otherwise, few patches dominate the attention. \textbf{Observations:}  1) ViT variants demonstrated sparse attention, while knowledge from convolution (\eg{}, DeiT-B-Distill) can largely smooth the attention weights, supported by both histogram and examples; 2) ViT variants exhibited a query-irrelevant (also head-irrelevant) behavior, supported by top examples. \textbf{Solution:} we address above issues by directly extracting a global context and introduce it into convolution, refer to \S\ref{sec:from_attention_to_convolution}.
    }
    \label{fig:teaser}
    \vspace{-1mm}
\end{figure}
In the past few years, Vision Transformers~\cite{vit,deit} has dominated various visual tasks in the computer vision community. 
Although the costs (\textit{i.e.}, parameters, and computations) are generally high, Vision Transformers are more likely to model better spatial relations and scale better with large models and datasets compared with the conventional convolutions~\cite{li2021benchmarking}. A common belief is that these gratifying virtues are credited to the self-attention mechanism. Nevertheless, this plausible conjecture has been challenged recently. Studies~\cite{wightman2021resnet,ding2022scaling, liu2022convnet} show that a ConvNet trained with strong training recipes can also achieve competitive or even higher performance, indicating that \textit{a deep investigation of the spatial modeling methods is worth further exploring}.

\begin{figure*}
    \centering
    \includegraphics[width=0.98\linewidth]{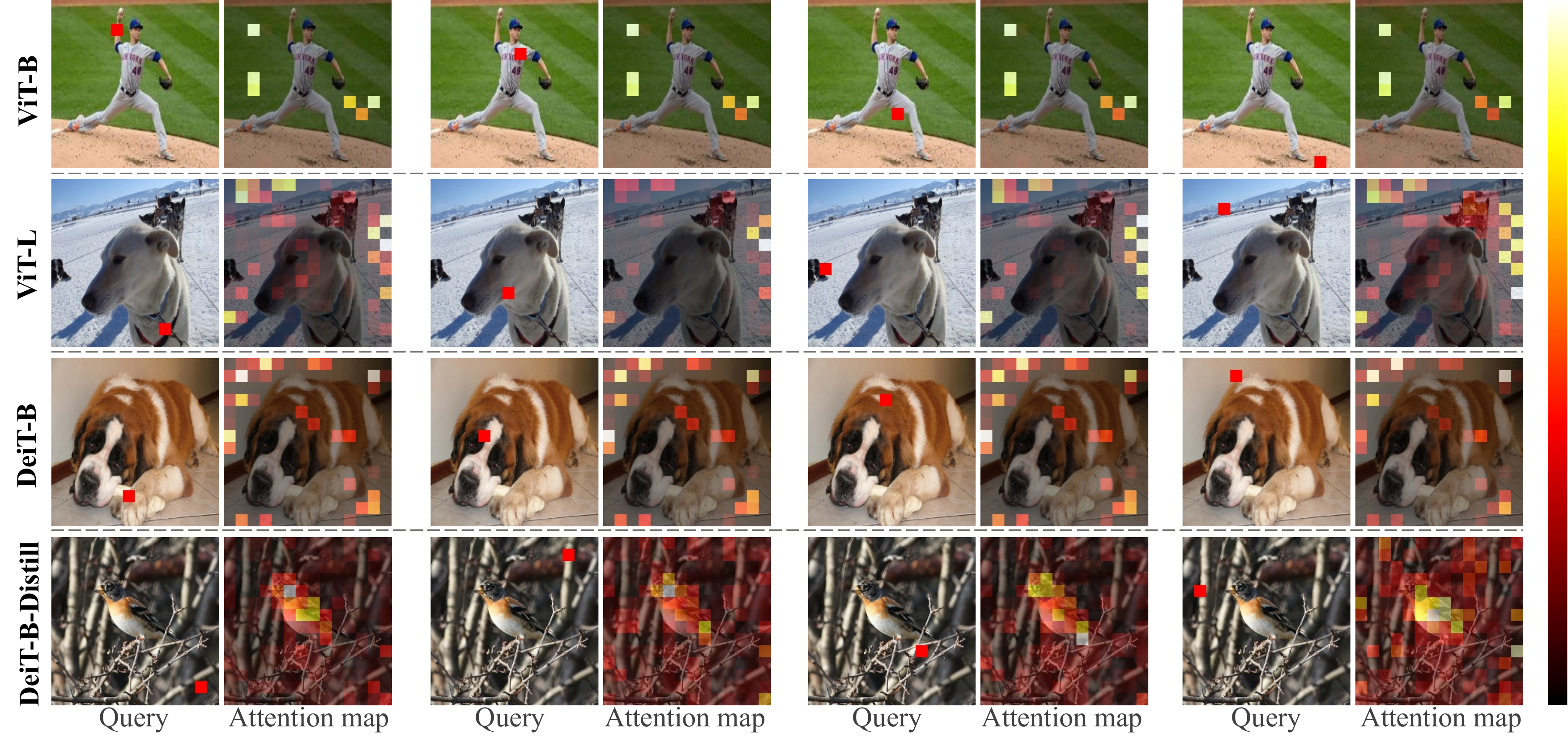}
    \vspace{-3mm}
    \caption{Attention map visualizations of Vision Transformers~\cite{vit,deit}. For each pair, we show the \textit{query point} and its corresponding \textit{attention map} (of last block and last head). All pre-trained models are downloaded from TIMM~\cite{rw2019timm}. The right color bar identifies the value of normalized attention maps. Surprisingly, \textcolor{black}{\textbf{the attention maps are almost the same, regardless of the query points}}. Moreover, \textcolor{black}{\textbf{the attention maps are intrinsically sparse if no knowledge from ConvNets is introduced}}. Results from different models indicate that \textbf{these phenomena are common in Vision Transformers}. See supplementary for more examples and \S~\ref{sec:introduction} for a more detailed analysis.  }
    \label{fig:motivation}
    \vspace{-3mm}
\end{figure*}
To verify what has been learned by self-attention in Vision Transformers, we take a close look at the attention maps of the deep layers in several representative Vision Transformers, as shown in Fig.~\ref{fig:motivation}. Astonishingly, all of these variants present a query-irrelevant behavior that reveals nearly consistent contexts in the global scope. This observation is a departure from the design philosophy of self-attention mechanism, indicating that a global context may be concealed behind the attention mechanism.
Meanwhile, we notice that the attention weight are considerably sparse, as shown in ViT and DeiT-B. By distilling the knowledge from ConvNets to Vision Transformer like DeiT-B-Distill, the attention map gets considerably smoother and concentrates more on objects. This phenomenon suggests that combining convolution and self-attention may lead to gratifying results. Notice that the above two observations are not just limited to particular images, a statistical analysis on ImageNet-1K dataset and fair comparisons shown in Fig.~\ref{fig:teaser} also confirm the pervasiveness.

Motivated by aforementioned findings, we propose \textbf{F}ully \textbf{C}onvolutional \textbf{Vi}sion \textbf{T}ransformer (\ie{}, FCViT) in this paper. 
Starting from the observation in Fig.~\ref{fig:motivation}, we progressively loosen the formulation of self-attention and abstract a global context that describes the global-range visual concepts. The global context is further dynamically introduced to local convolutional operations, making the efficient fusion of short- and long-range dependencies feasible. Intrinsically,
FCViT is a pure ConvNet but a Transformer-alike model that inherits the advantages of both Vision Transformer and ConvNet. 
Extensive experiments show that our FCViT consistently and significantly outperforms other methods with comparable costs. Remarkably, our FCViT achieves 80.9\% top-1 accuracy on ImageNet-1K~\cite{deng2009imagenet} using only 14M  parameters, outperforming ResT-Lite~\cite{zhang2021rest} and PoolFormer-S12 by \textbf{3.7\%}. Compared with state-of-the-art models like ConvNeXt~\cite{liu2022convnet}, we still present better results with even fewer parameters. FCViT also exhibits an excellent generalization ability. When transferred to downstream vision tasks like object detection and semantic segmentation, our FCViT consistently exhibits promising performance.

\vspace{-0.1cm}
\section{Related Work}
\vspace{-0.1cm}

\paragraph{Vision Transformers Dominate.} Originating from natural language processing, Transformer~\cite{vaswani2017attention} targets to dynamically build mutual relationships for each token pair in a global scope. Motivated by the successes in language, tentative efforts have been made toward migrating Transformer to the vision community. The pioneering work ViT~\cite{vit} has emerged as a promising approach for directly processing images using Transformer. Given an input image, ViT first tokenizes the input to non-overlapped patches and extracts token features via a stack of isotropic Transformer blocks. With an adequate training scheme, DeiT~\cite{deit} circumvents 
the problem of requiring large datasets. Since then, various Vision Transformer variants have been springing up~\cite{liu2021swin,yuan2021tokens}. Besides the aforementioned works, recent efforts~\cite{wang2021pyramid,yang2022focal, li2021localvit} mainly leverage the inductive biases from ConvNets to improve the performance. In this work, we push this trend further by connecting convolution and attention in FCViT. FCViT is a departure from standard Transformers, considering the fully convolutional operations; nevertheless, it unambiguously inherits the framework of the transformer architecture~\cite{yu2021metaformer} and enjoys the global receptive field.

\vspace{-4mm}
\paragraph{ConvNets Strike Back.} The necessity of multi-head self-attention in Transformers has been challenged, from language~\cite{JainW19,dong2021attention} to vision community~\cite{yu2021metaformer,zhai2021attention,tang2021sparse}. To put it another way, these critical investigations motivate lots of researchers to gush into the revitalization of ConvNets.
A systematic study is ConvNeXt~\cite{liu2022convnet}, which reexamines the design spaces of a ConvNet by gradually modifying ResNet~\cite{he2016deep} toward the standard vision Transformer~\cite{vit} (\ie{}, ViT). Astonishingly, ConvNeXt demonstrates that the resulting ConvNets compete favorably with Transformers in terms of both accuracy and scalability. Similar phenomena can also be observed in other ConvNets, like~\cite{bello2021revisiting, ding2022scaling}. 
Redesigning the architectures \textit{a la} the design philosophy of Transformers, ConvNets once again showed dominance in various tasks. This paper pushes the envelope further by inherently integrating virtues of ViTs with a fully convolutional network, which is where FCViT singularity lies.

\vspace{-4mm}
\paragraph{Convolution Meets Attention.} Another line of work on the topic of visual backbones learns to marriage the merits of both Transformers and ConvNets. That is, considering self-attention and convolution simultaneously in one token-mixer (mixing the spatial information). Dai \etal{}~\cite{dai2021coatnet} find that depthwise convolution and self-attention can be naturally unified in a token-mixer module. By doing so, CoAtNet~\cite{dai2021coatnet} effectively achieve better generalization and capacity. Similarly, CvT~\cite{wu2021cvt} introduces convolutional token embedding and convolutional projection into Vision Transformers and empirically presents better performances. Works in~\cite{guo2021cmt, chen2021mobile} bridge convolution module and attention module in a hybrid fashion. \textit{Different from the aforementioned methods that connect convolution and self-attention in an explicit manner, without intrinsic dedicated designs involved, our FCViT subtly bridges global context and local structure in one unified operation.} Besides, tentative efforts have been made toward exploring the relationships between convolution and attention.  
In this vein, our analysis is similar to those taken by~\cite{cordonnier2020relationship,han2022connection} and allows us to further improve the performance.

\section{From Attention to Convolution}
\label{sec:from_attention_to_convolution}
By revisiting self-attention and convolution, we first bridge self-attention and convolution in one unified operation. A close look at the attention maps in ViTs also demonstrates the necessity of integrating convolution and attention. Motivated by this, a simple yet effective Transformer-alike architecture is proposed to verify our findings.   Fig.~\ref{fig:framework} intuitively shows one building block of FCViT.

\subsection{A Close Look at Self-Attention and Convolution}
\subsubsection{Self-Attention and Convolution}
As the main contribution in Transformers, self-attention effectively captures the long-distance dependencies and dynamically aggregates input features according to the query patch. Formally, given an input feature map $\mathbf{X}\in\mathbb{R}^{d\times n}$, where $d$ is the embedding dimension, and $n$ indicates the patch number, the self-attention mechanism adaptively aggregates global information for each patch by 
\begin{equation}
    \label{eq:self_attention}
    \centering
    {
    \small
    \begin{split}
    \:\ y_i &= \sum_{j=1}^{n}w\left(q_i, k_j\right)v_j, \\
    \mathrm{s.t.,} \:\ w\left(q_i, k_j\right) &= \mathrm{softmax}\left(q_i^\top k_j\right)=\frac{ \mathrm{exp}\left(q_i^\top k_j\right)}{\sum_{l=1}^n \mathrm{exp}\left(q_i^\top k_l\right)},
    \end{split}
    }
\end{equation}
where $q_i=\mathbf{W}_q x_i$, $k_i=\mathbf{W}_k x_i$, and $v_i=\mathbf{W}_v x_i$ are different embeddings from $\mathbf{X}$; $i$, $j$, and $l$ index a patch. For brevity, we ignore the positional encoding and dimensional scalar $\sqrt{d}$ in Eq.~\ref{eq:self_attention}.

Differently, the convolution formula can be written as:
\begin{equation}
    \label{eq:convolution}
    y_i = \sum_{j\in \Omega}w_j x_j,
\end{equation}
where $\Omega$ indicates a local receptive field (\eg{}, a $3\times3$ kernel).

Comparing the formulations, two distinct differences between attention and convolution would be:

    $\bullet$ \textit{The dynamic of weights aggregation:}  self-attention dynamically aggregates the values $v_j$ using $w\left(q_i,k_j\right)$, which is based on the input features $x_i$. On the contrary, convolution utilizes fixed weights $w_j$ to fuse features, and the weights are shared across patches.
    \\
    $\bullet$ \textit{The size of receptive field:} self-attention can model long-distance dependencies among patches directly, resulting in a global receptive field. Limited by the computations, convolution performs aggregation functions in a local region, like $3\times 3$ and $7 \times 7$. A sliding window pattern (with a small step size) guarantees the information flow among regions. By stacking multiple layers, convolution can progressively enlarge the receptive field.

\begin{figure*}
    \centering 
    \includegraphics[width=0.8\linewidth]{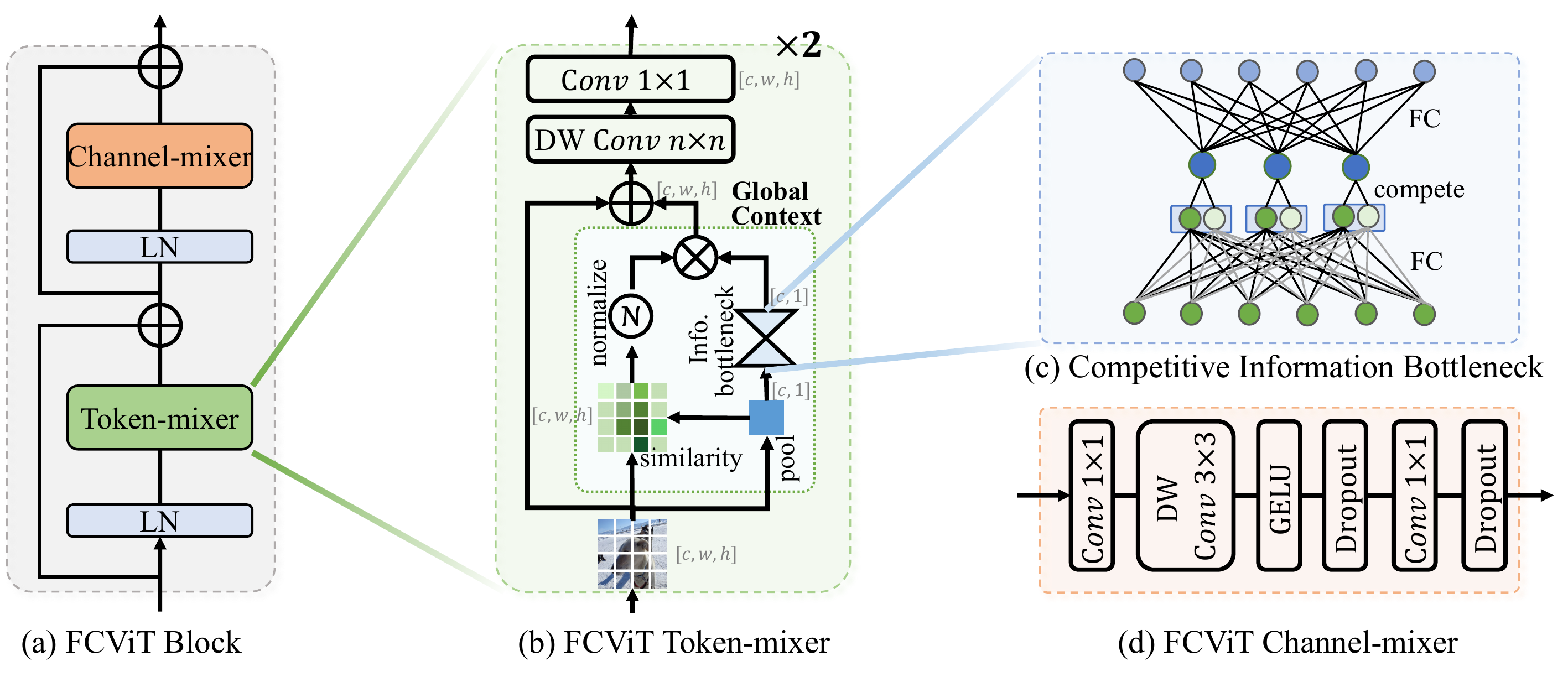}
    \vspace{-0.2cm}
    \caption{Illustration of an FCViT block. FCViT considers the block as a combination of token-mixer and channel-mixer. In the token-mixer, we dynamically integrate the global context with input tokens by the token-global similarity. A depth-wise convolution is employed to fuse local information. To improve the generalization of the global context, we introduce a competition-driven information bottleneck structure. Overall FCViT configurations are presented in the supplementary.}
    \vspace{-0.2cm}
    \label{fig:framework}
\end{figure*}

\subsubsection{Attention Map Visualization}
Some conventional wisdom leans to credit the success of Vision Transformers to the attention mechanism, considering the two differences discussed above. We acknowledge the advantages of a dynamic scheme and large receptive field; however, a close look into the details is still worth further exploring. To this end, we visualize the attention maps of various Vision Transformers, including ViT-B, ViT-L, DeiT-B, and DeiT-B-Distill. Results in Fig.~\ref{fig:motivation} present some interesting phenomena: 

\begin{itemize}
    \item \textbf{Observation 1:} the attention maps consistently show a query-irrelevant (and even head-irrelevant) behavior. Visually, the attention maps $w\left(q_i, k_j\right)$ appear to be nearly identical for each testing model and image, regardless of the query patch $q_i$. This is a departure from the design philosophy of self-attention that each patch should exhibit a distinct attention map.  
    \vspace{-3mm}
    \item \textbf{Observation 2:} the attention weights (see ViT-B, ViT-L, and DeiT-B) are relatively sparse, indicating that only several patches dominate the attention. By introducing the knowledge from convolution, the attention weights (see DeiT-B-Distill) are largely smoothed, and the performance is significantly improved as well (83.4\% of DeiT-B-Distill \textit{vs.} 81.8\% of DeiT-B top-1 accuracy on ImageNet-1K validation set). 
\end{itemize}

The aforementioned two observations suggest that: \textbf{1)} a query-irrelevant global context may be sufficient to work well for vision tasks; \textbf{2)} combining the knowledge from both self-attention and convolution can yield gratifying results.

\subsubsection{From Self-Attention to Convolution}
Inspired by the aforementioned observations, we revisit self-attention and connect it to convolution. Given the Eq.~\ref{eq:self_attention}, we first remove the querying patches as implied by observation 1. The resulting simplified attention can be written as  
\begin{equation}
    \label{eq:simple_attention}
    \begin{split}
    y = \sum_{j=1}^{n}w\left(k_j\right)v_j &=\sum_{j=1}^{n}\frac{ \mathrm{exp}\left(k_j\right)}{\sum_{l=1}^n \mathrm{exp}\left(k_l\right)}v_j \\
      &=\mathcal{N}(\mathbf{K})\mathbf{V} = gc,
    \end{split}
\end{equation}
where $\mathcal{N}(\cdot)$ indicates a normalization function, like softmax. Since Eq.~\ref{eq:simple_attention} is purely based on the global input features $\mathbf{X}$ and models the global information, we denote it as global context (\ie{}, $gc\in \mathbb{R}^d$) for simplicity. Note that the objective of normalization function $\mathcal{N}(\cdot)$ is to generate a weight for $v_i$. Hence, it can be generalized to other forms besides softmax, like average or sigmoid function. 

Furthermore, as indicated by observation 2, we are encouraged to combine the attention mechanism and convolution into a unified formulation. A simple implementation that bridges Eq.~\ref{eq:simple_attention} and Eq.~\ref{eq:convolution} is
\begin{equation}
    \label{eq:simple_atten+conv}
     y_i = \sum_{j\in \Omega}w_j (x_j+gc).
\end{equation}

Conceptually, Eq.~\ref{eq:simple_atten+conv} introduces a global context into each patch and leverages a convolution to aggregate local information consequently. With Eq.~\ref{eq:simple_atten+conv}, we efficiently introduce the global information into the local patches, avoiding the combination of two individual operations. Meanwhile, by employing the global context implemented in Eq~\ref{eq:simple_attention}, we significantly reduce the computational complexity of self-attention. Also, Eq.~\ref{eq:simple_atten+conv} can be easily implemented by a convolution layer, making our model concise yet effective.
Next, we instantiate our FCViT based on the analysis above.

\subsection{Fully Convolutional Vision Transformer}

\subsubsection{General Framework}
Our FCViT follows the framework of MetaFormer~\cite{yu2021metaformer}, which adopts hierarchical architecture with 4 stages that progressively reduce the spatial size. Given an input image, we first utilize overlapping patch embedding to tokenize images by linear mapping. In each stage, a series of isotropic FCViT blocks are utilized to extract features. 
FCViT block involves two independent modules, the token-mixer and channel-mixer (as well as residual connections and Layer Normalization), as shown in Fig.~\ref{fig:framework}. In the end, a classifier is employed to generate the classification logits. Varying the number of blocks and the channel number, we instantiate FCViT by FCViT-Tiny, FCViT-B12, and FCViT-B24, \etc{}. Detailed configurations can be found in the supplementary. Next, we describe the detailed designs of our FCViT.

\subsubsection{Enhanced Global Context in Token-Mixer}
Following Eq.~\ref{eq:simple_atten+conv}, our token-mixer can be implemented by
\begin{equation}
    \label{eq:token-mixer}
    \mathbf{Y} = \mathrm{conv}_{1} \left(
    \mathrm{conv}_{k} 
    \left(
     \mathbf{X}+gc
    \right)
    \right) 
    = \mathrm{conv}\left(
     \mathbf{X}+gc
    \right),
\end{equation}
where $\mathrm{conv}_1$ is a point-wise convolution and $\mathrm{conv}_k$ is a depth-wise convolution with kernel size of~$k$.  For convenience, we use $\mathrm{conv}$ to denote the combination of the two convolutions. We repeat the operation in Eq.~\ref{eq:token-mixer} twice to achieve the best model size and accuracy trade-off. Specifically, we normalize the global context by average-pooling instead of softmax for simplicity. That is:
\begin{equation}
    \label{eq:gap_gc}
    gc = \sum_{i=1}^n \frac{\mathbf{W}_vx_i}{n}
    =\mathbf{W}_v\sum_{i=1}^n \frac{x_i}{n}.
\end{equation}

\paragraph{Dynamic Global Context.}
Directly fusing the global context $gc\in \mathbb{R}^d$ with input feature $\mathbf{X}\in \mathbb{R}^{d\times w\times h}$  may lead to limited improvements since $gc$ is broadcasted equally for each patch in $\mathbf{X}$. In other words, Eq.~\ref{eq:token-mixer} can be rewritten as
\begin{equation}
    \label{eq:token-mixer-rewritten}
    \begin{split}
        \mathbf{Y}
        &= \mathrm{conv}\left( \mathbf{X}+\mathbf{W}_v\sum_{i=1}^n \frac{x_i}{n}\right) \\
        &= \mathrm{conv}\left(\mathbf{X}\right)+\mathbf{W}_{conv}^{\top}\mathbf{W}_v\sum_{i=1}^n \frac{x_i}{n}
    ,
    \end{split}
\end{equation}
where $\mathbf{W}_{conv}^{\top}$ is the weights in the convolution layer. \textit{Eq.~\ref{eq:token-mixer-rewritten} explicitly demonstrates that directly fusing $gc$ with $\mathbf{X}$ is equal to fusing $gc$ outsides the convolution.} Hence, we expect a dynamic fusion scheme that fuses the global context with bias based on the input $\mathbf{X}$.

We circumvent this problem by promoting token-global similarity. For clarity, we denote $\sum_{i=1}^n \frac{x_i}{n}$ as $\overline{\mathbf{X}}$. The similarity score $\mathbf{S}\in\mathbb{R}^{w\times h}$ is given by $\mathbf{S} = \mathbf{X} \overline{\mathbf{X}}$, which calculates the relations between each patch $x_i$ and the global average pooling $\overline{\mathbf{X}}$. We re-scale the similarity and update $gc$ by
\begin{equation}
    \label{eq:updated_gc}
    gc' = \left(\alpha \frac{\mathbf{S}-\mu_{\mathbf{S}}}{\sigma_{\mathbf{S}} +\epsilon}+\beta\right)gc = \mathbf{S}'gc,
\end{equation}
where $\alpha$ and $\beta$ are learnable scalars; $\mu_{\mathbf{S}}$ and $\sigma_{\mathbf{S}}$ are mean and standard deviation of $\mathbf{S}$; $\epsilon=1e^{-5}$ is for numerical stability. We use $\mathbf{S}'$ and $gc'$ to present normalized $\mathbf{S}$ and updated $gc$, respectively. By doing so, each patch dynamically integrates the global context according to the similarity.

\paragraph{Multi-Group Similarity.} Inspired by multi-head attention, we further extend our token-global similarity to a multi-group fashion in pursuit of better diversity. We first divide  $gc$, $\mathbf{X}$, and $\overline{\mathbf{X}}$ into $g$ groups along the channel dimension, respectively. We then perform the operation of Eq.~\ref{eq:updated_gc} for each group and merge the outputs via concatenation. That is, 
$
    \label{eq:group_version}
     gc' = \mathrm{concat}\left([\mathbf{S}_1'gc_1, \mathbf{S}_2'gc_2, \cdots, \mathbf{S}_g'gc_g]\right).
$
Different from the multi-head attention that maps all channels into multiple heads, we leverage the group operation to reduce computational overheads. Ablation study in Fig.~\ref{fig:ablation_vis_compare} visually shows the differences between our multi-group similarity and multi-head attention.

\paragraph{Competitive Information Bottleneck.} 
We next introduce a competitive information bottleneck to further improve the generalization ability of $gc$ and better describe the global context. We first squeeze $\overline{\mathbf{X}}$ by $\mathbf{W}_{s1}\in \mathbb{R}^{\frac{d}{r}\times d}$ and $\mathbf{W}_{s2}\in \mathbb{R}^{\frac{d}{r}\times d}$, respectively. Then, the two squeezed vectors compete to generate a bottleneck representation by maxout~\cite{goodfellow2013maxout}. Lastly, we recover the representation to original dimension by $\mathbf{W}_{r}\in\mathbb{R}^{d\times \frac{d}{r}}$. Our competitive information bottleneck can be written as
$
    gc = \mathbf{W}_{r}\mathrm{maxout}\left(\mathbf{W}_{s1}\overline{\mathbf{X}}, \mathbf{W}_{s2}\overline{\mathbf{X}}\right).
$
By default, we set $r$ to 8. Then, the multi-group dynamic $gc$ can be applied consequently. While the design is similar to SENet~\cite{hu2018squeeze}, we are in pursuit of competitive information and parameters reduction; no attention design is involved.
In spite of the fact that it is not essential to our FCViT, the competitive information bottleneck can significantly reduce parameters and consistently boost performance.

\subsubsection{Conducive Implementation Details}
While the token-mixer is the core of FCViT, some detailed designs are also instrumental. Different from the non-overlapping tokenization in ViT~\cite{vit} and ConvNeXt~\cite{liu2022convnet}, we consider the overlapped patch embedding as presented in ~\cite{wang2021pvtv2}, where each patch has small overlapping to ease the communication among patches. Also, a depth-wise convolution is introduced between the two point-wise convolution layers, as shown in Fig.~\ref{fig:framework}.

\section{Experiments}
We validate FCViT on ImageNet-1K~\cite{deng2009imagenet}, MS COCO~\cite{lin2014microsoft}, and ADE20K~\cite{zhou2017scene} datasets. We first demonstrate the effectiveness of FCViT on ImageNet-1K classification task. Extensive ablation studies provide a close look at the internal operations. Then, we transfer pre-trained models to object detection, instance segmentation, and semantic segmentation tasks to examine the generalization ability of FCViT.
\begin{table}[!t]\centering
{\small
\begin{tabular}{l|cccc}
    \Xhline{3\arrayrulewidth}
	\whiteshape{} Method & \makecell{ Param.\\(M)} & \makecell{ FLOPs\\(G)} & \makecell{ Top-1\\(\%)}& \makecell{ Speed\\(im/s)}  \\
	\hline
	\attshape{} PVTv2-B0 ~\cite{wang2021pvtv2} & 3.4 & 0.6 & 70.5  &- \\
	\attshape{} T2T-ViT-7 ~\cite{yuan2021tokens} & 4.3 & 1.1 & 71.7&- \\ 
	\attshape{} DeiT-Tiny/16~\cite{deit} & 5.7 & 1.3 & 72.2   &767.07\\
    \attshape{} TNT-Ti~\cite{han2021transformer} & 6.1 & 1.4 & 73.9 &- \\
    \rowcolor{RowColor}
    \convshape{}FCViT-Tiny $_{\mathrm{(ours)}}$ &4.6  & 0.8&\textbf{74.9}  & 759.79\\

	\hline
	\attshape{} PVT-Tiny~\cite{wang2021pyramid} & 13.2 & 1.9 &75.1  &- \\
	\attshape{} ResT-Lite~\cite{zhang2021rest} &10.49 & 1.4& 77.2  &-\\
	\attshape{} SOFT-Tiny~\cite{SOFT} &13.0 & 1.9& 79.3  &-\\
	\othershape{} Pool-S12~\cite{yu2021metaformer} & 11.9  & 2.0 & 77.2  &764.03\\ 
	\convshape{} ResNet18~\cite{he2016deep} & 12 & 1.8 & 69.8  & 926.73  \\
	\rowcolor{RowColor} \convshape{}
    FCViT-B12 $_{\mathrm{(ours)}}$ &14  &2.5 &\textbf{80.9} &771.56 \\

	\hline
	\attshape{} DeiT-Small/16~\cite{deit}  & 22.1 & 4.6 & 79.8  & 762.19 \\
	\attshape{} PVT-Small~\cite{wang2021pyramid}  & 24.5 & 3.8 & 79.8  &724.52\\
	\attshape{} Swin-T~\cite{liu2021swin} & 28.3 & 4.5 & 81.3   &758.84\\
	\attshape{} ResT-Base~\cite{zhang2021rest} &30.28 & 4.3& 81.6  &-\\
	\mlpshape{} ResMLP-24~\cite{touvron2021resmlp} & 30.0 & 6.0 & 79.4  &756.88 \\
	\mlpshape{} AS-MLP-T~\cite{Lian_2021_ASMLP} & 28 & 4.4 & 81.3 &- \\
	\othershape{} Pool-S24~\cite{yu2021metaformer} & 21.4  & 3.6 & 80.3 & 763.77 \\ 
	\convattshape{} CoAtNet-0~\cite{dai2021coatnet} &25 &4.2 &81.6 &-\\ %
	\convattshape{} CvT-13~\cite{wu2021cvt} &20 &4.5 &81.6 &- \\
	\convattshape{} Conformer-Ti~\cite{peng2021conformer}&23.5 &5.2 &81.3 &- \\
	\convshape{} ResNet50~\cite{he2016deep}   &26 &4.1 &80.4  & 770.34 \\
	\convshape{} ConvNeXt-T~\cite{liu2022convnet} & 28.6 & 4.5 & 82.1 & 747.27\\  
	\convshape{} PatchConv-S60~\cite{touvron2021augmenting} &25.2 & 4.0 & 82.1 &- \\
	\convshape{} Focal-T(SRF)~\cite{yang2022focal} & 28.4 & 4.4 & 82.1&708.46 \\
	\convshape{} Focal-T(LRF)~\cite{yang2022focal} & 28.6 & 4.5 & 82.3 &725.04 \\
    \rowcolor{RowColor} \convshape{}
    FCViT-B24 $_{\mathrm{(ours)}}$ &25.7  &4.7 &\textbf{82.5} &706.97 \\

\Xhline{3\arrayrulewidth}
\end{tabular}
}
\caption{
Comparison with SOTA backbones on ImageNet-1K benchmark. Throughput (images / s) is measured on a single A100 GPU with a batch size of 128. All models are trained and evaluated on 224×224 resolution. We use dots with different colors to present different types of token-mixer, \textcolor{attcolor}{attention-based}, \textcolor{convcolor}{convolution-based}, \textcolor{mlpcolor}{MLP-based}, \textcolor{attconvcolor}{Att\&Conv.-based}, and \textcolor{othercolor}{other} token-mixers. The best results are marked in \textbf{bold}. For more results of larger models, please see the supplementary.
}
\label{tab:imagenetcls}
\vspace{-1mm}
\end{table}

\subsection{Image Classification on ImageNet-1K}
\paragraph{Experimental Settings.} We train FCViT models on the ImageNet-1K training set (with around 1.3M images) and evaluate upon the validation set. We follow the common training recipe in ~\cite{dai2021coatnet,rw2019timm,deit,yu2021metaformer}. 
All our models are trained for 310 epochs using AdamW~\cite{loshchilov2018decoupled} with a momentum of 0.9 and a weight decay of 0.05. The learning rate is initialized to 0.001 and adjusted by cosine schedular~\cite{loshchilov2016sgdr}. By default, the models are trained on 8 A100 GPUs with a mini-batch size of 128 (1024 in total). Similar to previous works~\cite{guo2022visual,deit}, we employ Exponential Moving Average (EMA) to improve the training.
Results are reported in Table~\ref{tab:imagenetcls}.

\noindent\textbf{Performance Analysis.} Empirically, our FCViT outperforms related work by a clear margin. FCViT-Tiny outperforms DeiT-Tiny by \textbf{2.7\%} (74.9\% \textit{vs.} 72.2\%) top-1 accuracy, using much fewer parameters and FLOPs. 
When increasing the model size, FCViT consistently achieves a leading performance. FCViT-B12 achieves 80.9\% accuracy with 14M parameters, outperforming ResT-Lite and PoolFormer-S12 by \textbf{3.7\%}. Comparing with the state-of-the-art model ConvNeXt, we still present an improvement of 0.4\% (82.5\% \textit{vs.} 82.1\%) accuracy, with fewer parameters but similar FLOPs.

\subsection{Isotropic FCViT}
We also investigate the compatibility of FCViT block with isotropic design (\eg{}, DeiT~\cite{deit}, ResMLP~\cite{touvron2021resmlp}), in which no down-sampling block is introduced and the feature map resolution is constant across all depths. To build Isotropic FCViT,
we "patchify" input images via a stride-$p$ $p\times p$ convolution, with $p=16$ by default, like ViT. Then, isotropic FCViT blocks are stacked to extract features. "256/12" indicates the input channel of each block is 256 and we use 12 FCViT blocks to build the network (same meaning for "384/16"). Isotropic FCViTs are trained with the same settings as before. From the resulting in Table~\ref{tab:isotropic}, we observe Iso. FCViTs perform better than related isotropic architectures, indicating the effectiveness of our design.

\begin{table}[]
    \centering
    {\small
    \begin{tabular}{l|ccc}
    \Xhline{3\arrayrulewidth}
    Model & Param.(M) & FLOPs.(G)& Top-1(\%)  \\
    \hline
    
        DeiT-Ti~\cite{deit} &5.7&1.3 &72.2  \\
        \rowcolor{RowColor}Iso.-FCViT-256/12 &8.2 &1.4 &\textbf{75.0}\\
        \hline
        DeiT-S~\cite{deit}  &22.1&4.6 &79.8 \\
        ResMLP-S24~\cite{touvron2021resmlp} &30 &6.0 &79.4 \\
        MLP-Mixer-B/16~\cite{tolstikhin2021mlp} &59.9 &12.6 &76.4 \\
        ConvNeXt-S({\small iso.})~\cite{liu2022convnet}&22.0 &4.3 &79.7 \\
        ConvMixer-768/32 &21.1 &20.9 & 80.2\\
        \rowcolor{RowColor}Iso.-FCViT-384/16 &23.2 &4.0 &\textbf{80.3}\\

    \Xhline{3\arrayrulewidth}
    \end{tabular}
    }
    \caption{Comparison between Isotropic FCViT and other isotropic architectures on ImageNet-1K benchmark.}
    \label{tab:isotropic}
\end{table}

\subsection{Ablation Study}
We next conduct extensive ablation studies to better understand and evaluate the utility of FCViT. Detailed analyses demonstrate the contribution of introducing the dynamic global context to local regions. Component ablations further disentangle the effectiveness of each module. In this subsection, we conduct all experiments based on FCViT-T, and we train all models with a mini-batch size of 256.

\begin{figure*}%
    \centering
    \subfloat[\centering FCViT-B12 token-global similarity.\label{fig:ablation_vis_compare_left}]{{
    \includegraphics[width=0.338\linewidth]{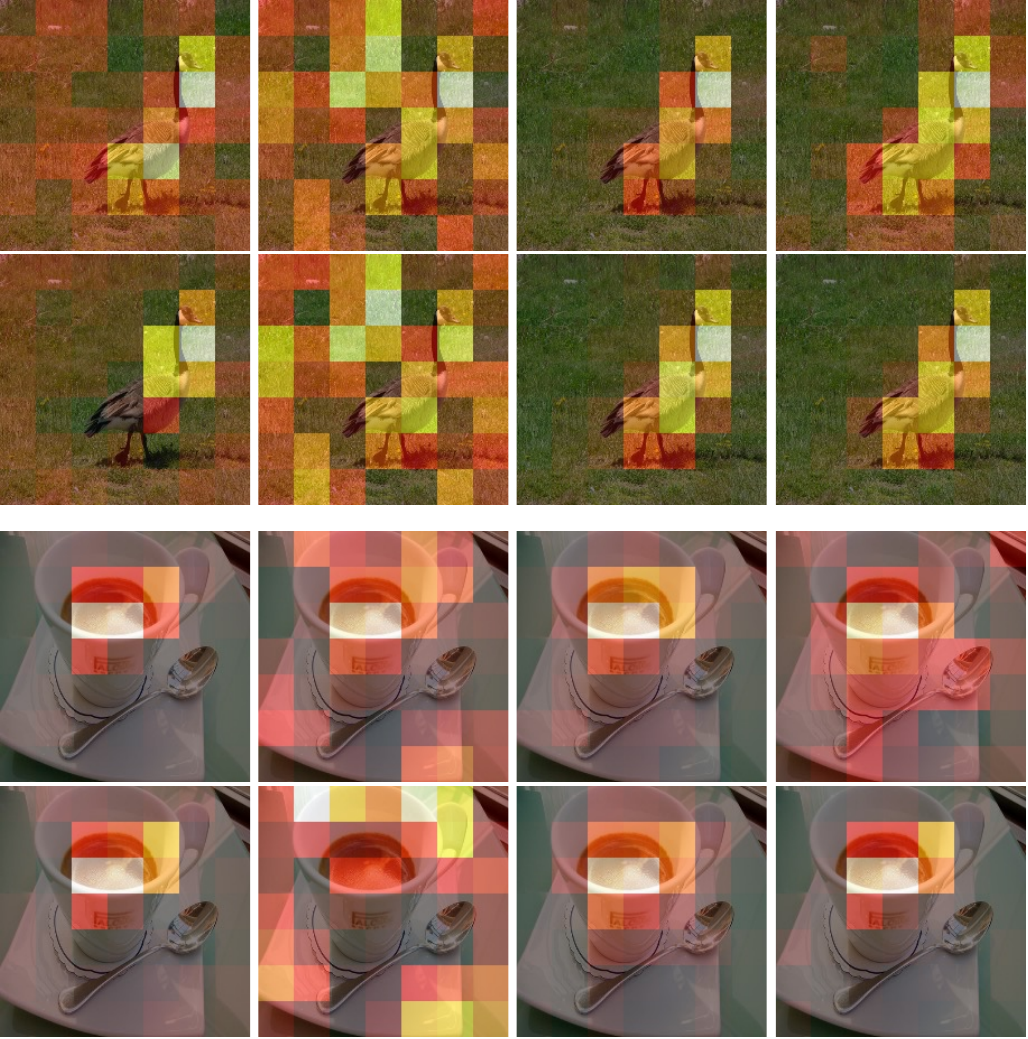} }}%
    \qquad
    \subfloat[\centering ViT-B  query point and self-attention maps.\label{fig:ablation_vis_compare_right}]{{
    \includegraphics[width=0.6\linewidth]{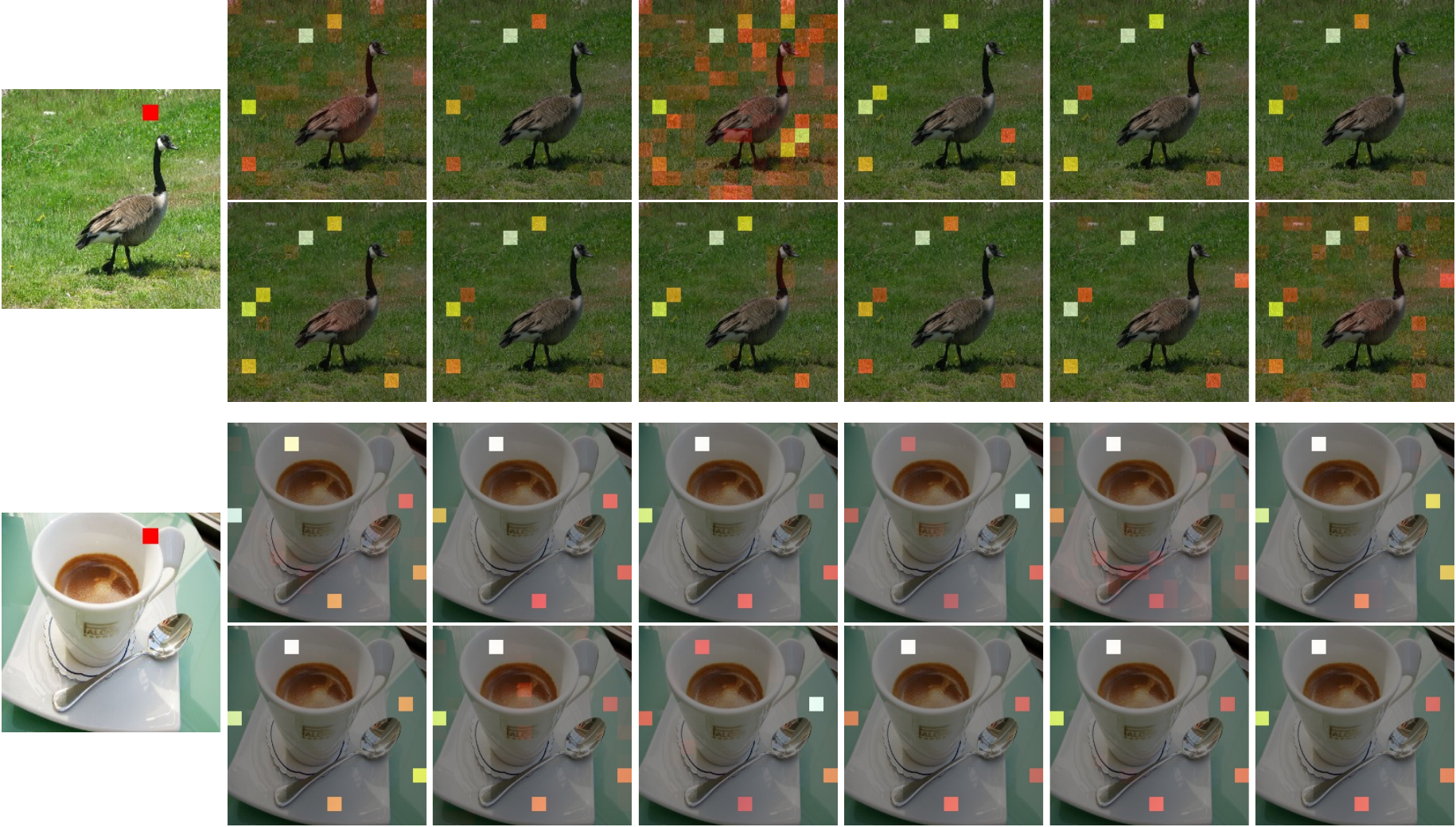} }}%
    \vspace{-2mm}
    \caption{Visual comparisons of FCViT-B12 similarity and ViT-B attention map. We plot all the outputs of the last block for the two models (8 groups for FCViT and 12 heads for ViT). Compared to ViT, the results indicate that: \textbf{1)}, our FCViT focuses more on the objects; \textbf{2)} FCViT presents more diversities than multi-head attention, whose attention maps from different heads are nearly the same.}%
    \label{fig:ablation_vis_compare}%
\end{figure*}

\begin{table}[!h]
    \centering
    \begin{tabular}{l|cccccccc}
        Size& 3 & 5 & 7&8 & 9 &11 & 13  \\
        \Xhline{3\arrayrulewidth}
        Acc.& 73.2 &73.2& 73.4& 73.4 &73.2 & \textbf{73.5}& 73.2
    \end{tabular}
    \vspace{-9mm}
\end{table}
\paragraph{Kernel Size.} 
We vary the kernel size of convolutions in the token-mixer from 3 to 13. Different from previous work~\cite{liu2022convnet}, we did not observe a linear correlation between kernel size and performance, and the performance gap is relatively small, ranging from 73.2\% to 73.5\%. One possible reason would be that we explicitly introduced the global context to the local tokens, which already enlarges the receptive field to a global range. Hence, the influence of kernel size is largely diluted. By default, we set the value to 11.

\begin{table}
    \centering

    \begin{tabular}{ccc|ccc}
        \Xhline{3\arrayrulewidth}
        \makecell{GC}& \makecell{ Dy.\\GC.}  & \makecell{ Comp.\\Info.}  & params & FLOPs & top-1 (\%) \\
        \hline
        \xmark &\xmark &\xmark  &4.2M  &0.8G & 72.8\textcolor{cyan}{\tiny{$\uparrow$0.0}}\\
        \cmark &\xmark &\xmark  &4.4M  &0.8G & 73.0\textcolor{cyan}{\tiny{$\uparrow$0.2}} \\
        \cmark &\cmark &\xmark  &4.4M  &0.8G & 73.5\textcolor{cyan}{\tiny{$\uparrow$0.7}} \\
        \cmark &\cmark &\cmark &4.5M  &0.8G & 73.7\textcolor{cyan}{\tiny{$\uparrow$0.9}} \\
        \Xhline{3\arrayrulewidth}
    \end{tabular}
    \caption{Ablation of Global Context. We individually ablate the global context, dynamic scheme, and competitive bottleneck.}
    \label{tab:compoent_ablation}
    \vspace{-2.5mm}
\end{table}
\paragraph{Global Context.} 
The kernel contribution of FCViT is to introduce the global context to local tokens, converting self-attention to local convolution. Here, we evaluate the necessity of our global context. We first build a plain FCViT version as our baseline, where the global context is removed. Then, we integrate the dynamic version and competitive information bottleneck progressively. Table~\ref{tab:compoent_ablation} shows the effectiveness of each component. Without Global Context, a plain FCViT architecture reaches 72.8\% top-1 accuracy, which is still a leading result in the first block of Table~\ref{tab:imagenetcls}. By introducing the global context, we slightly improve the plain network by 0.2\%. When integrating the global context as presented in Eq.~\ref{eq:updated_gc}, the performance is increased to 73.5\%, indicating the effectiveness of the dynamic scheme. Remarkably, the computation and parameter overheads are negligible, leading to almost no additional inference time. When introducing the competitive information bottleneck, we further improve the performance to 73.7\%, exhibiting 0.9\% improvements over the plain network without global context. In the following experiments, we will consider the global context with the dynamic scheme and competitive information bottleneck by default.

\begin{table}[!h]
    \centering
    \begin{tabular}{l|ccccc}
        Group& 1 &4 &8 &16& 32  \\
        \Xhline{3\arrayrulewidth}
        Acc.& 73.2 & 73.4\footnotesize{($\uparrow$)} & \textbf{73.5} &  73.3 &  73.3
    \end{tabular}
    \vspace{-3mm}
\end{table}
\noindent\textbf{Group Number for Token-Global Similarity.} As introduced, we intentionally copy the design philosophy of multi-head attention to emphasize the similarity in different groups. The table on the right reveals the benefits of multi-group similarity. When the group number is increased to 8, we achieve the best result (0.3\% improvements).
Note that only memory operation is introduced for multi-group similarity (computational complexity and operational redundancy are almost the same), and the running speed is not increased. Empirically, we set the group number to 8 in all other experiments.

\begin{table*}[]
    \centering
     
     \vspace{-0.2cm}
    
    \begin{tabular}{l|c|lcc|lcc}
        \Xhline{3\arrayrulewidth}
        backbone & Params    &AP$^{\mathrm{box}}$ &AP$^{\mathrm{box}}_{50}$  & AP$^{\mathrm{box}}_{75}$& AP$^{\mathrm{mask}}$& AP$^{\mathrm{mask}}_{50}$ & AP$^{\mathrm{mask}}_{75}$ \\
        \hline
        ResNet-18~\cite{he2016deep} &31.2M         &34.0  &54.0 & 36.7       &31.2  &51.0 &32.7\\
        PoolFormer-S12~\cite{yu2021metaformer} &31.6M         &37.3  &59.0 & 40.1       &34.6  &55.8 &36.9\\
        PVT-Tiny~\cite{wang2021pyramid} &32.9M        &36.7  &59.2 &39.3        &35.1  &56.7 &37.3\\
        \rowcolor{RowColor}FCViT-B12 {\small (ours)}& 34.3M & \textbf{42.3} &\textbf{64.2} &\textbf{46.2}& 
        \textbf{38.6} &\textbf{61.1} &\textbf{41.3}\\
        \hline
        ResNet-50~\cite{he2016deep} &44.2M       &38.0  &58.6 &41.4        &34.4  &55.1 &36.7\\
        PoolFormer-S24~\cite{yu2021metaformer} &41.0M     &40.1  &62.2 &43.4       & 37.0 &59.1 &39.6\\
        PVT-Small~\cite{wang2021pyramid} &44.1M         &40.4  & 62.9&43.8        & 37.8 &60.1 &40.3\\
        Swin-Tiny~\cite{liu2021swin,han2022vision}  &47.8M         &42.2 &64.6 &46.2 &39.1 &61.6 &42.0\\
        Swin-Tiny~\cite{liu2021swin,yang2022focal}  &47.8M         &43.7 &\textbf{66.6} &47.7 &39.8 &\textbf{63.3} &42.7\\
        \rowcolor{RowColor}FCViT-B24 {\small (ours)}& 43.1M & \textbf{44.1} &65.4 &\textbf{48.4}& 
        \textbf{39.9} &62.4 &\textbf{42.7}\\
        \Xhline{3\arrayrulewidth}
    \end{tabular}
    \vspace{-2mm}
    \caption{COCO object detection and instance segmentation results using Mask-RCNN (1$\times$).}
    \label{tab:detection}

\end{table*}

\subsection{Multi-Group Token-Global Similarity \textit{vs.} Multi-Head Self-Attention.} Similar to the multi-head self-attention mechanism, we consider the token-global similarity in a grouped-channel fashion. In Fig.~\ref{fig:ablation_vis_compare_left}, we showcase the similarity scores of FCViT-B12 for all 8 groups in the last block. Similarly, we present the attention maps of ViT-B for all 12 heads in Fig.~\ref{fig:ablation_vis_compare_right}. We randomly fix the query point for all validating images since Fig.~\ref{fig:motivation} demonstrated that the attention map of the last block is nearly query-irrelevant. 

The visualization results exhibit some interesting insights. 
\textbf{1)} Besides the query-irrelevance, we notice that the  attention map of ViT is head-irrelevant as well, as shown in Fig.~\ref{fig:ablation_vis_compare_right}. This is a departure from the design philosophy of multi-head attention mechanism, where different heads suppose to exhibit different attention. Differently, our FCViT presents diverse similarities among the groups. This phenomenon can be explained by the formulations. In multi-head self-attention, all channels are 
employed to map to multiple heads. Without constraints (\eg{}, loss functions), the heads are hard to present meaningful diversities, especially when the model converges in the last layers. As a comparison, our channel-group similarity calculates the similarity for a group of channels individually, making each group self-contained and preventing the similarities collapsed. 
\textbf{2)} Our FCViT focuses more on the objects, while ViT cannot exhibit such a desirable property. Among all testing examples and groups, FCViT explicitly emphasizes the tokens within the objects. Recall that the similarity is calculated between local tokens and the global context. This result highlights the validity of our global context. Meanwhile, by dynamically fusing the global context, our FCViT further enhances the representational ability of local tokens in turn.

\begin{table}
    \centering
    
    \begin{tabular}{l|cc}
        \Xhline{3\arrayrulewidth}
        Backbone & Params & mIoU(\%)  \\
        \hline
        ResNet18~\cite{he2016deep}       & 15.5M  & 32.9 \\
        PVT-Tiny~\cite{wang2021pyramid} & 17.0M & 35.7\\
        PoolFormer-S12~\cite{yu2021metaformer}& 15.7M &37.2\\
        \rowcolor{RowColor}FCViT-B12 {\small (ours)}& 17.8M  &\textbf{43.3}\\
        \hline
        ResNet50~\cite{he2016deep} & 28.5M  &36.7\\
        PVT-Small~\cite{wang2021pyramid}& 28.2M& 39.8\\
        Poolformer-S24~\cite{yu2021metaformer}&23.2M &40.3\\
        Twins-PCPVT-S~\cite{chu2021twins}&28.4M &44.3\\
        Twins-SVT-S~\cite{chu2021twins}& 28.3M & 42.6 \\
        \rowcolor{RowColor}FCViT-B24 {\small (ours)}& 25.3M & \textbf{45.5}\\
        \Xhline{3\arrayrulewidth}
        
    \end{tabular}
    \vspace{-1mm}
    \caption{Semantic segmentation performance  of different backbones with Semantic FPN on the ADE20K validation set.}
    \vspace{-3mm}
    \label{tab:ade20k}
\end{table}
\subsection{Object Detection and Instance Segmentation}
We next probe the transferability of FCViT on downstream tasks, including object detection and instance segmentation. We conduct experiments on MS COCO 2017 benchmark~\cite{lin2014microsoft}. We train and evaluate Mask R-CNN ~\cite{he2017mask} with FCViT backbone initialized with classification pre-trained weights. For a fair comparison, we follow the settings in PVT~\cite{wang2021pyramid} and PoolFormer~\cite{yu2021metaformer} that adopts the 1$\times$ training schedule (\ie{}, 12 epochs) based on the MMDetection~\cite{mmdetection} framework. 
Results in Table~\ref{tab:detection} demonstrate that our FCViT significantly outperforms related work by a clear margin. Particularly, our FCViT brings \textbf{5.0 points of mAP$^\mathrm{box}$  and 4.0 points of mAP$^\mathrm{mask}$} against PoolFormer-S12 at comparable settings. Even compared with larger models (parameters in backbones are doubled) like PoolFormer-S24 and ResNet-50, our FCViT-Tiny still exhibits significantly better results, showing gratifying transferability. 

\subsection{Semantic Segmentation on ADE20K}

We also evaluate our FCViT equipped with Semantic FPN~\cite{kirillov2019panoptic} on ADE20K~\cite{zhou2017scene} dataset for the semantic segmentation task. 
ADE20K contains 150 semantic categories and consists of 20k, 2k, and 3k images for training, validation, and testing, respectively.
We use the ImageNet-pretrained backbone model taken from Table~\ref{tab:imagenetcls}. Following~\cite{yu2021metaformer}, we train all our models for 40k iterations with a batch size of 32. All our models are trained using AdamW optimizer with an initial learning rate of 2x10-4. We decay the learning rate by a polynomial decay schedule with a power of 0.9.

The results compared with previous work are presented in Table~\ref{tab:ade20k}. Empirically, our FCViT achieves promising performance on the semantic segmentation task. Compared with PoolFormer-S12, FCViT-B12 improves the performance by 6.1\% mIoU using a similar number of parameters. Compared with Twins-PCPVT-S, FCViT-B24 improves the performance by 1.2\% mIoU using even fewer parameters. In line with the results in object detection and instance segmentation tasks, FCViT also demonstrated promising transferability for the semantic segmentation task.

\vspace{-0.1cm}

\section{Conclusion}
In this paper, we first take a close look at two foundational token-mixers, attention and convolution. Observations of sparse and query-irrelevant attention maps motivate us to connect attention with convolution by abstracting a global context and dynamically fusing it with local tokens. The resulting model, Fully Convolutional Vision Transformer (FCViT), explicitly embraces the advantages of both convolution and attention, and exhibits promising efficiency and performance. The multi-group token-global similarity and competitive information bottleneck further boost the representational ability of our method. Experiments on multiple tasks validate the utility of FCViT and confirm our analysis. We hope that our empirical observations and FCViT design can bring new insights to the vision community.

{\small
\bibliographystyle{ieee_fullname}
\bibliography{egbib}
}
\newpage
\appendix

\twocolumn[{%
\renewcommand\twocolumn[1][]{#1}%
\maketitle
\begin{center}
  \centering
  \captionsetup{type=figure}
  \includegraphics[width=0.9\linewidth]{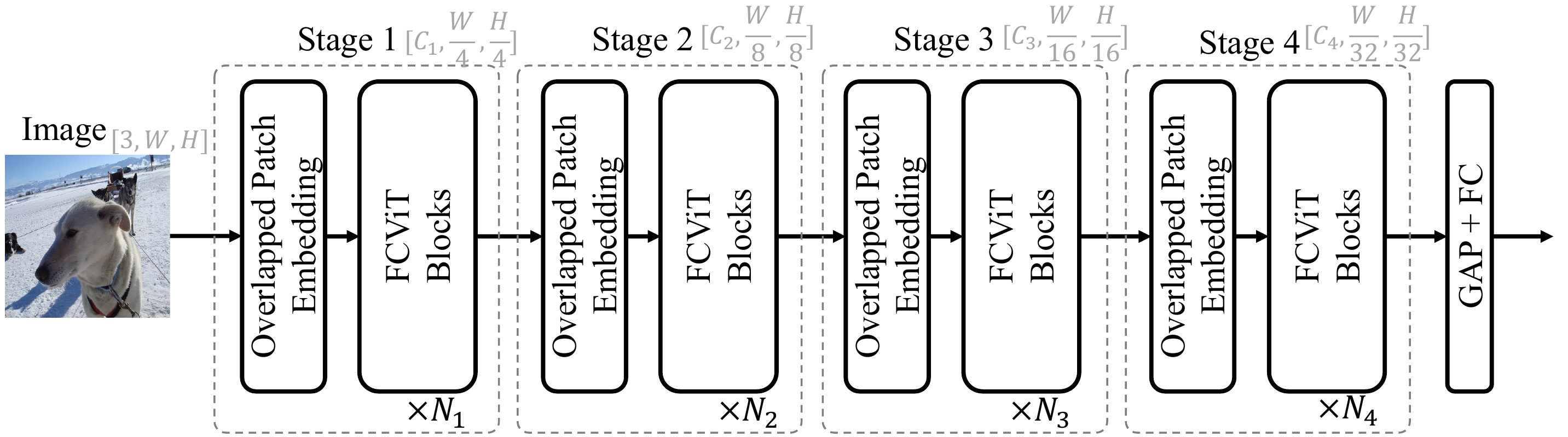}
  \captionof{figure}{
    The overall architecture of FCViT. Given an input image, FCViT considers 4 stages to process and extract features. In each stage, one overlapped patch embedding is employed to reduce the spatial size, and a stack of FCViT blocks is employed to model feature relations.
  }
  \label{fig:architecture}
\end{center}%
}]

\section{FCViT Architecture \& Configurations}

\subsection{FCViT Architecture}
We first describe the high-level architecture overview of our FCViT. Following hierarchical Transformers~\cite{liu2021swin} and classical ConvNets~\cite{he2016deep}, our FCViT includes four stages to reduce the spatial size gradually. Specifically, we reduce the spatial size by a factor of 4 in the first stage and a factor of 2 for the rest. In the end, a classifier (which is combined by a global averaging pooling and a fully connected layer) is employed to give the classification logits. 

Next, we exhaustively present the details of overlapped patch embedding. Different from the non-overlapping patch embedding used in the original ViT and ConvNeXt, our overlapped patch embedding introduces appropriate overlapping between neighbor patches. In our implementation, we achieve this by a convolution operation, where the stride size is smaller than the kernel size. For example, the kernel size is set to 7, and the step size is set to 4 in the first stage. As a result, half of the pixels in a patch are overlapped with neighbor patches. To facilitate the training, we add Layer Normalization after the convolution operation, which is consistent with previous work.

\subsection{FCViT Configurations}
The detailed configurations of our FCViT variants are shown in Table~\ref{tab:configutations}. We instantiate four variants of FCViT for different model capacities. Empirically, we set the MLP-ratio in the channel-mixer (FFN) to 8 for the first and second stages. For the blocks number, we follows the design in ~\cite{liu2022convnet,yu2021metaformer}: the number is set to $n$, $n$, $3n$, and $n$ for each stage.

\begin{table*}

{\footnotesize
\resizebox{.99\textwidth}{!}{%
\begin{tabular}{c|c|c|c|c|c|c}
\Xhline{3\arrayrulewidth}
 Stage&Size & Layer & FCViT-Tiny & FCViT-B12 & FCViT-B24 & FCViT-B48 \\
 \hline
\multirow{2}{*}{S1} & 56x56 & 
\makecell{Patch\\Embed.} & $\left[\makecell{\mathrm{patch\_size}=7 \\ \mathrm{stride}=4 \\\mathrm{dim}=32} \right]$&  
$\left[\makecell{\mathrm{patch\_size}=7 \\ \mathrm{stride}=4 \\\mathrm{dim}=64} \right]$
&  
$\left[\makecell{\mathrm{patch\_size}=7 \\ \mathrm{stride}=4 \\\mathrm{dim}=64} \right]$
& 
$\left[\makecell{\mathrm{patch\_size}=7 \\ \mathrm{stride}=4 \\\mathrm{dim}=64} \right]$\\
\cline{2-7}
 \rule{0pt}{5ex} & 56x56 & \makecell{FCViT\\Blocks} 
 &  $\left[\makecell{\mathrm{mlp\_r.}=8 \\\mathrm{dim}=32} \right]\times 3$
 &  $\left[\makecell{\mathrm{mlp\_r.}=8 \\\mathrm{dim}=64} \right]\times 2$
 &  $\left[\makecell{\mathrm{mlp\_r.}=8 \\\mathrm{dim}=64} \right]\times 4$
 &  $\left[\makecell{\mathrm{mlp\_r.}=8 \\\mathrm{dim}=64} \right]\times 8$
 \\
 \hline
\multirow{2}{*}{S2} & 28x28 & \makecell{Patch\\Embed.} 
&  $\left[\makecell{\mathrm{patch\_size}=3 \\ \mathrm{stride}=2 \\\mathrm{dim}=64} \right]$
&  $\left[\makecell{\mathrm{patch\_size}=3 \\ \mathrm{stride}=2 \\\mathrm{dim}=128} \right]$
&  $\left[\makecell{\mathrm{patch\_size}=3 \\ \mathrm{stride}=2 \\\mathrm{dim}=128} \right]$
&  $\left[\makecell{\mathrm{patch\_size}=3 \\ \mathrm{stride}=2 \\\mathrm{dim}=128} \right]$
\\
\cline{2-7}
 \rule{0pt}{5ex} & 28x28 & \makecell{FCViT\\Blocks} &  $\left[\makecell{\mathrm{mlp\_r.}=8 \\\mathrm{dim}=64} \right]\times 3$
 &  $\left[\makecell{\mathrm{mlp\_r.}=8 \\\mathrm{dim}=128} \right]\times 2$
 &  $\left[\makecell{\mathrm{mlp\_r.}=8 \\\mathrm{dim}=128} \right]\times 4$
 &  $\left[\makecell{\mathrm{mlp\_r.}=8 \\\mathrm{dim}=128} \right]\times 8$  \\
 \hline
\multirow{2}{*}{S3} & 14x14 & \makecell{Patch\\Embed.} 
&  $\left[\makecell{\mathrm{patch\_size}=3 \\ \mathrm{stride}=2 \\\mathrm{dim}=160} \right]$
&  $\left[\makecell{\mathrm{patch\_size}=3 \\ \mathrm{stride}=2 \\\mathrm{dim}=320} \right]$
&  $\left[\makecell{\mathrm{patch\_size}=3 \\ \mathrm{stride}=2 \\\mathrm{dim}=320} \right]$
&  $\left[\makecell{\mathrm{patch\_size}=3 \\ \mathrm{stride}=2 \\\mathrm{dim}=320} \right]$
\\
\cline{2-7}
 \rule{0pt}{5ex}& 14x14 & \makecell{FCViT\\Blocks} & $\left[\makecell{\mathrm{mlp\_r.}=4 \\\mathrm{dim}=160} \right]\times 5$
 &  $\left[\makecell{\mathrm{mlp\_r.}=4 \\\mathrm{dim}=320} \right]\times 6$
 &  $\left[\makecell{\mathrm{mlp\_r.}=4 \\\mathrm{dim}=320} \right]\times 12$
 &  $\left[\makecell{\mathrm{mlp\_r.}=4 \\\mathrm{dim}=320} \right]\times 24$\\
 \hline
\multirow{2}{*}{S4} & 7x7 & \makecell{Patch\\Embed.} 
& $\left[\makecell{\mathrm{patch\_size}=3 \\ \mathrm{stride}=2 \\\mathrm{dim}=320} \right]$ 
& $\left[\makecell{\mathrm{patch\_size}=3 \\ \mathrm{stride}=2 \\\mathrm{dim}=512} \right]$ 
& $\left[\makecell{\mathrm{patch\_size}=3 \\ \mathrm{stride}=2 \\\mathrm{dim}=512} \right]$ 
& $\left[\makecell{\mathrm{patch\_size}=3 \\ \mathrm{stride}=2 \\\mathrm{dim}=512} \right]$ \\
\cline{2-7}
 \rule{0pt}{5ex}& 7x7 & \makecell{FCViT\\Blocks} & $\left[\makecell{\mathrm{mlp\_r.}=4 \\\mathrm{dim}=320} \right]\times 2$
 &  $\left[\makecell{\mathrm{mlp\_r.}=4 \\\mathrm{dim}=512} \right]\times 2$
 &  $\left[\makecell{\mathrm{mlp\_r.}=4 \\\mathrm{dim}=512} \right]\times 4$
 &  $\left[\makecell{\mathrm{mlp\_r.}=4 \\\mathrm{dim}=512} \right]\times 8$\\
\Xhline{3\arrayrulewidth}
\end{tabular}
}
\caption{Model configurations for our FCViT. Based on the framework of MetaFormer, we introduce four configurations FCViT-Tiny, FCViT-B12, FCViT-B24, and FCViT-B48, with different model scales and capacities.}
\label{tab:configutations}
}
\end{table*}

\section{More Experiments}
We present more experiments, including large model comparisons and more ablation studies.

\begin{table}[!t]\centering
\begin{tabular}{l|ccc}
    \Xhline{3\arrayrulewidth}
	\whiteshape{} Method & \makecell{ Param.\\(M)} & \makecell{ FLOPs.\\(G)} & \makecell{ Top-1.\\(\%)}  \\
	\hline
	\attshape{} T2T-ViT$_t$-19~\cite{yuan2021tokens} & 39.2 & 9.8 & 82.4  \\
	\attshape{} PVT-Medium~\cite{wang2021pyramid} & 44.2M & 6.7 & 81.2 \\
	
	\attshape{} Swin-S ~\cite{liu2021swin} & 49.6 & 8.7 & 83.0 \\
	
	\attshape{} PVTv2-B3 ~\cite{wang2021pvtv2} & 45.2 & 6.9 & 83.2 \\
	\attshape{} Focal-S~\cite{yang2021focal} & 51.1 & 9.1 & 83.5\\
	\attshape{} T2T-ViT$_t$-24~~\cite{yuan2021tokens} & 64.0 & 15.0 & 82.3 \\
	\mlpshape{} AS-MLP-S~\cite{Lian_2021_ASMLP} & 50 & 8.5 & 83.1\\ 
	\mlpshape{} Mixer-B/16~\cite{tolstikhin2021mlp} & 59.0 & 11.6 & 76.4\\
    \othershape{} PoolFormer-M36~\cite{yu2021metaformer}& 56 &9.1& 82.1 \\
    \othershape{} PoolFormer-M48~\cite{yu2021metaformer}& 73 &11.9& 82.5  \\
    \othershape{} S2-MLP-deep~\cite{yu2022s2} &51 &10.5 &80.7  \\
    \convattshape{} CoAtNet-1~\cite{dai2021coatnet} &42 &8.4 &83.3 \\
    \convattshape{} CvT-21~\cite{wu2021cvt} &32 &7.1 &82.5 \\
    \convattshape{} Conformer-S~\cite{peng2021conformer}&37.7 &10.6 &83.4  \\
    \convshape{} ResNet101~\cite{he2016deep}   &45 &7.9 &81.3  \\
    \convshape{} ConvNeXt-S ~\cite{liu2021swin} & 50.1 & 8.7 & 83.1 \\
    \rowcolor{RowColor} \convshape{}
    FCViT-B48$_{\mathrm{(ours)}}$ & 49.1 &9.2 & \textbf{83.6} \\
	\hline

\Xhline{3\arrayrulewidth}
\end{tabular}
\caption{Comparison with larger SOTA backbones on ImageNet-1k benchmark. 
The best results are marked in \textbf{bold}.
}
\vspace{0.1cm}
\label{tab:imagenet_cls_supp}
\end{table}

\begin{figure}
    \vspace{-4.5mm}
    \centering
    \includegraphics[width=0.9\linewidth]{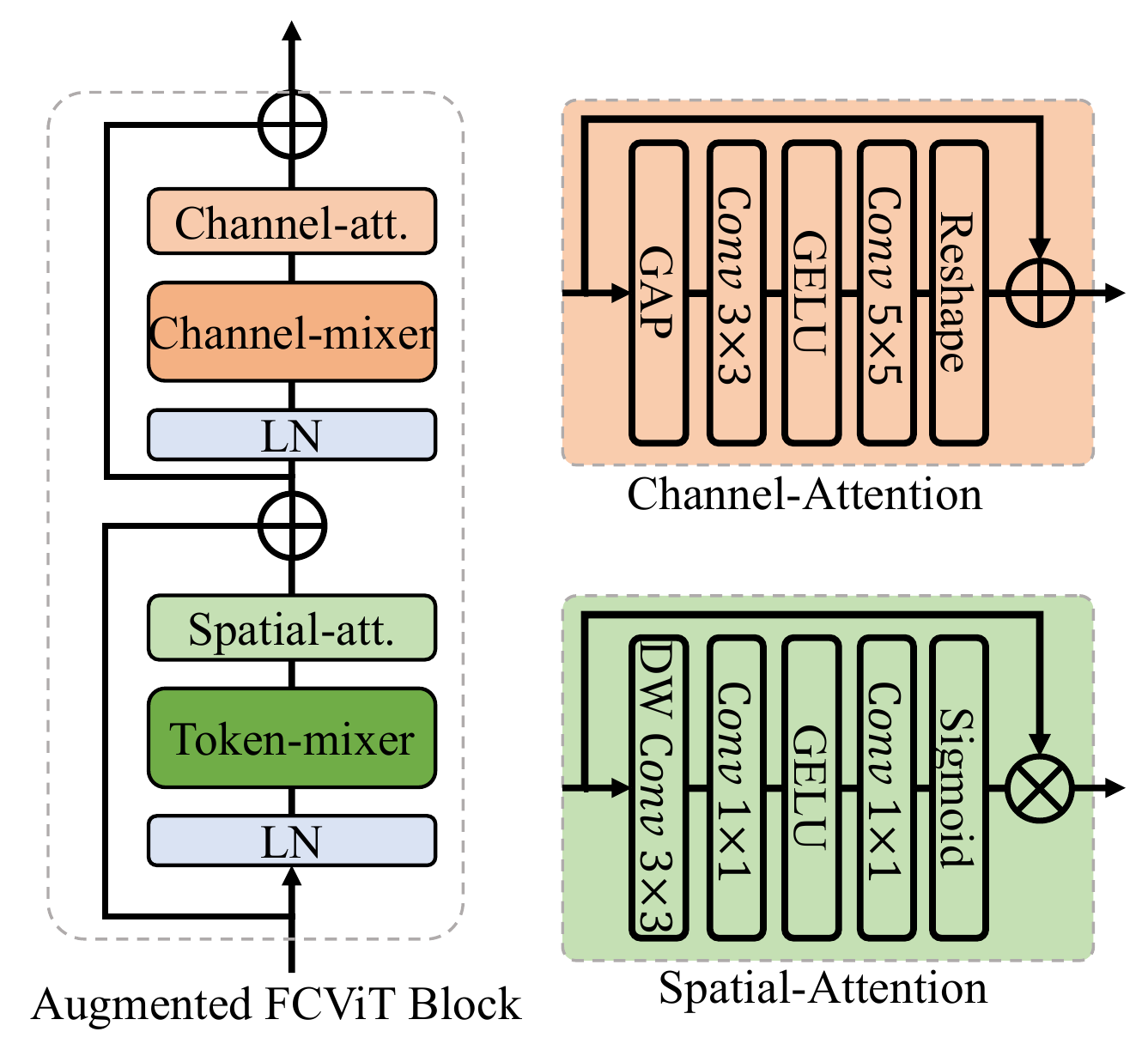}
    \vspace{-3mm}
    \caption{Attention module augmented FCViT block, and details in spatial- and channel-attention.}
    \vspace{-5mm}
    \label{fig:attention}
\end{figure}
\paragraph{Larger Model.} First, we introduce the performance of larger models as a supplementary. Clearly, our FCViT-B48 still dominates the classification performance with relatively few parameters for the larger models. This promising result indicates that our FCViT not only surpasses related work significantly with a small model capability, but also enjoys a good scalability.
\paragraph{Integration of Attention Module.} Some works introduce attention modules laterally to enhance the model with minimal additional overheads~\cite{hu2018squeeze}. We examine the effectiveness of combining attention module with our FCViT to verify if attention modules can boost our FCViT as well.

To achieve this, we first build two fully convolutional attention modules, the spatial-attention module and the channel-attention module. Then, we plug the two modules into token-mixer and channel-mixer, respectively. Fig.~
\ref{fig:attention} show the attention module augmented FCViT block. In spatial attention, we introduce two depth-wise convolutions and two point-wise convolutions to model the local spatial attention. In channel-attention, we first abstract the global feature via global average pooling, then two 1D convolutions are applied along the channel dimension to module the channel dependencies. We employ the summation for the channel-attention in our implementation. Note that both our spatial- and channel-attention are based on convolutional operation, aligning with our motivation and design philosophy. 

\begin{table}
    \centering
    
    \begin{tabular}{cc|cc}
        \Xhline{3\arrayrulewidth}
        Spatial-att.& Channel-att. &top-1 &top-5  \\
        \hline
        \xmark & \xmark &74.9 &92.6\\
        \cmark & \xmark &74.9 &92.6\\
        \xmark & \cmark &75.0 &92.6\\
        \cmark & \cmark &75.1 &92.5\\
        \Xhline{3\arrayrulewidth}
    \end{tabular}
   \caption{Ablation on attention modules.}
    \label{tab:attention_module}
\end{table}

\begin{figure*}
    \centering
    \includegraphics[width=0.95\linewidth]{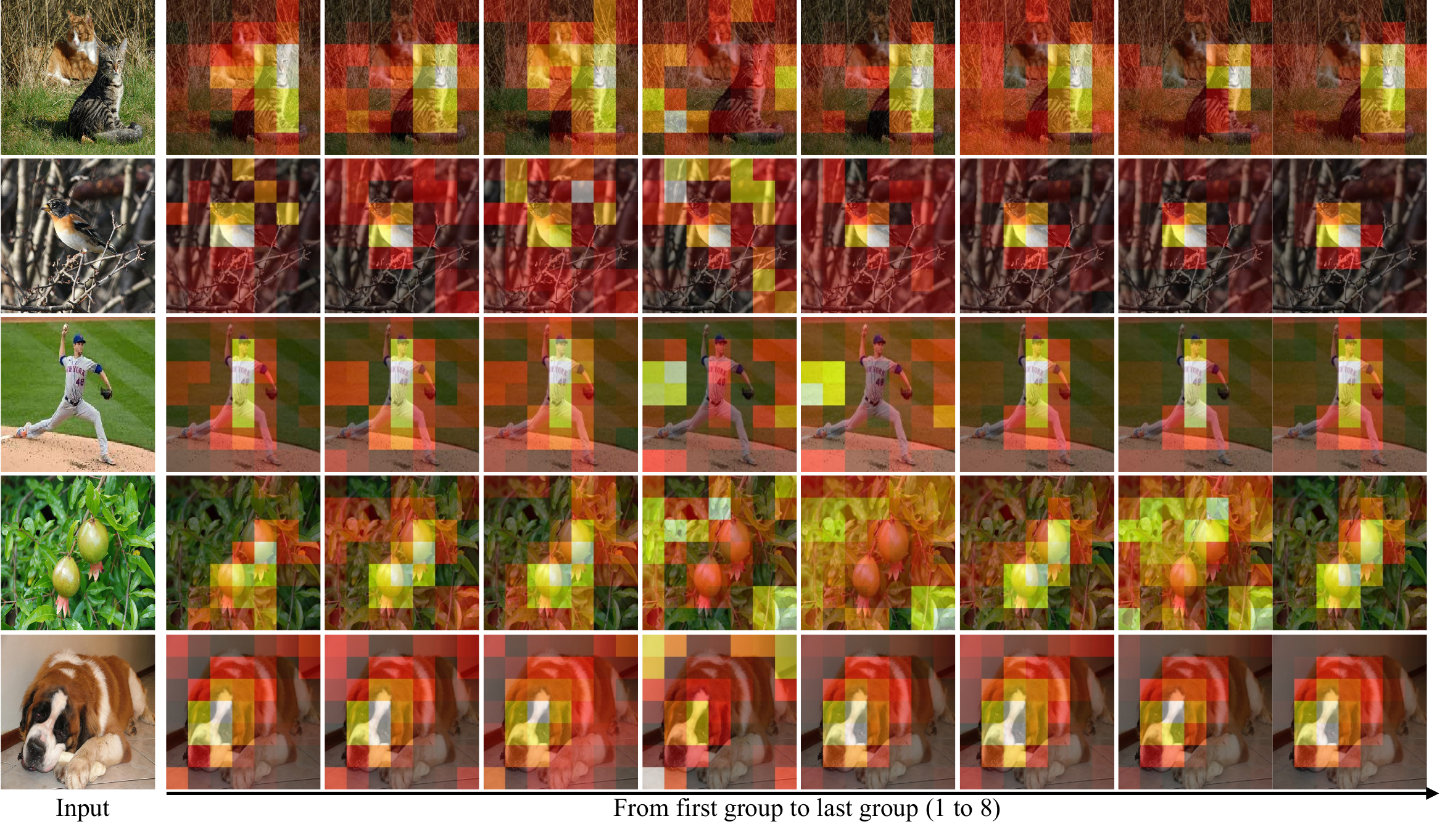}
    \vspace{-3mm}
    \caption{More token-global similarity visualization. Obviously, each group exhibits distinct similarities, and the similarities in all groups consistently emphasize the critical regions.}
    \label{fig:supp_sim}
\end{figure*}

Results in Tale~\ref{tab:attention_module} exhibit some interesting phenomena. Different from previous works~\cite{hu2018squeeze,woo2018cbam,hu2018gather}, additional attention modules improve our FCViT marginally, only 0.1\% to 0.2\% top-1 accuracy and no improvements on top-5 accuracy. The following two aspects can explain this: 1) our FCViT explicitly inherits both global- and local-range feature modeling abilities, making additional attention modules show limited improvements; 2) the training recipe further limits the contributions of attention modules. Unlike conventional ConvNets trained with 100 epochs with simple data augmentation methods, recent works train all models by 300 epochs and with more and better training strategies.   
Besides our tailored convolutional attention modules, SE module can also be employed as a channel-attention module in our method, but no improvements were observed. Considering the computation and parameter overheads, we use our channel-attention module, which introduces 10 parameters.
Considering the limited contribution of additional attention modules and strong baselines achieved by FCViT, we did not include convolutional attention modules in our architecture.

\paragraph{Repeated token-mixer.} As shown in the FCViT Token-mixer, we repeat the global context and depth-wise convolution by two times. By reducing this operation repetition time to one, the performance decreases 0.9\%. More repetitions would largely increase the computational overhead. Hence, we set the repetition to 2 in our implementation. 
\paragraph{Depth-wise Convolution in Channel-Mixer(FFN).} 
Results indicate that this simple modification can improve the performance by 1.2\% (73.7\% \textit{vs.} 74.9\%). Empirically, we set the kernel size to 3. Other values may lead to better performance, but not the key contributions to our work.

\paragraph{Removing each component in enhanced global context.}
 Our enhanced global context has three components: dynamic global context, multi-group similarity, and competitive information bottleneck. We remove each component individually and present the results in  Table~\ref{tab:remove_individual_component}. Clearly, each component can contribute to our FCViT. Combining them together, we achieve the best performance.

\begin{table}
    \centering
    {\small
    \begin{tabular}{cccc|c}
        \Xhline{3\arrayrulewidth}
        GC & Dynamic& Comp. Info & Group Sim. & top1 \\
        \cmark &\cmark &\cmark &\cmark & 73.7$ \textcolor{white}{_{(- 0.0)}}$ \\
        \xmark &\xmark &\xmark &\xmark & 72.8 $_{(\downarrow 0.9)}$ \\
        \cmark &\xmark &\cmark &\cmark & 73.2 $_{(\downarrow 0.5)}$\\
        \cmark &\cmark &\xmark &\cmark & 73.5 $_{(\downarrow 0.2)}$\\
        \cmark &\cmark &\cmark &\xmark & 73.3 $_{(\downarrow 0.4)}$ \\
        \Xhline{3\arrayrulewidth}
    \end{tabular}
    }
    \caption{Ablation of each component in enhanced global context.}
    \label{tab:remove_individual_component}
\end{table}

\begin{table}[]
    \centering
    \begin{tabular}{c|ccc|c}
         \Xhline{3\arrayrulewidth}
        Token-mixer&Param. &FLOPs &Blocks & Top-1 \\
        \hline
        Self-attention &4.8M & 0.8G & [3,3,5,2] & 74.7\\
        Conv-3x3     & 4.8M  & 0.9G &[3,3,9,2]& 74.6  \\
        Conv-11x11  &4.8M  &1.0G& [3,3,7,2] & 74.5 \\
        \rowcolor{RowColor}FCViT-Tiny  & 4.6M  & 0.8G & [3,3,5,2]& \textbf{74.9} \\
        \Xhline{3\arrayrulewidth}
    \end{tabular}
    \caption{Replacing Enhanced Global Context to Convolution or Attention. See the description for detail modifications.}
    \label{tab:replace}
\end{table}

\begin{figure*}
    \centering
    \includegraphics[width=0.88\linewidth]{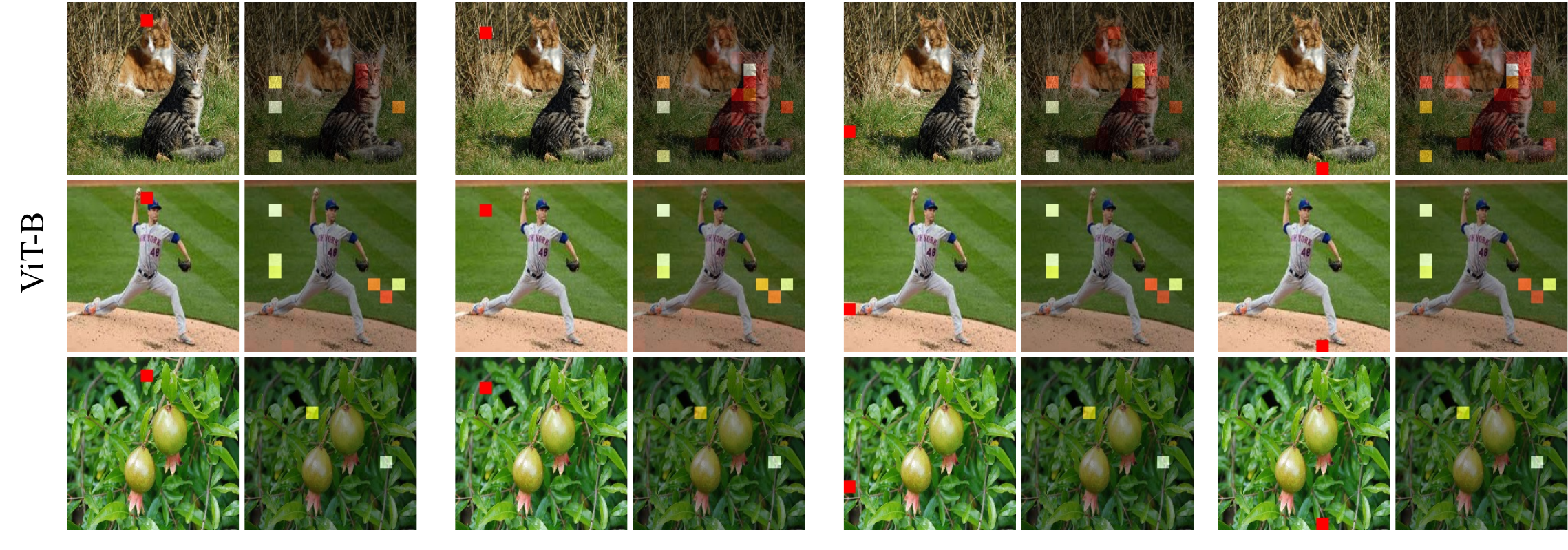}
    \includegraphics[width=0.88\linewidth]{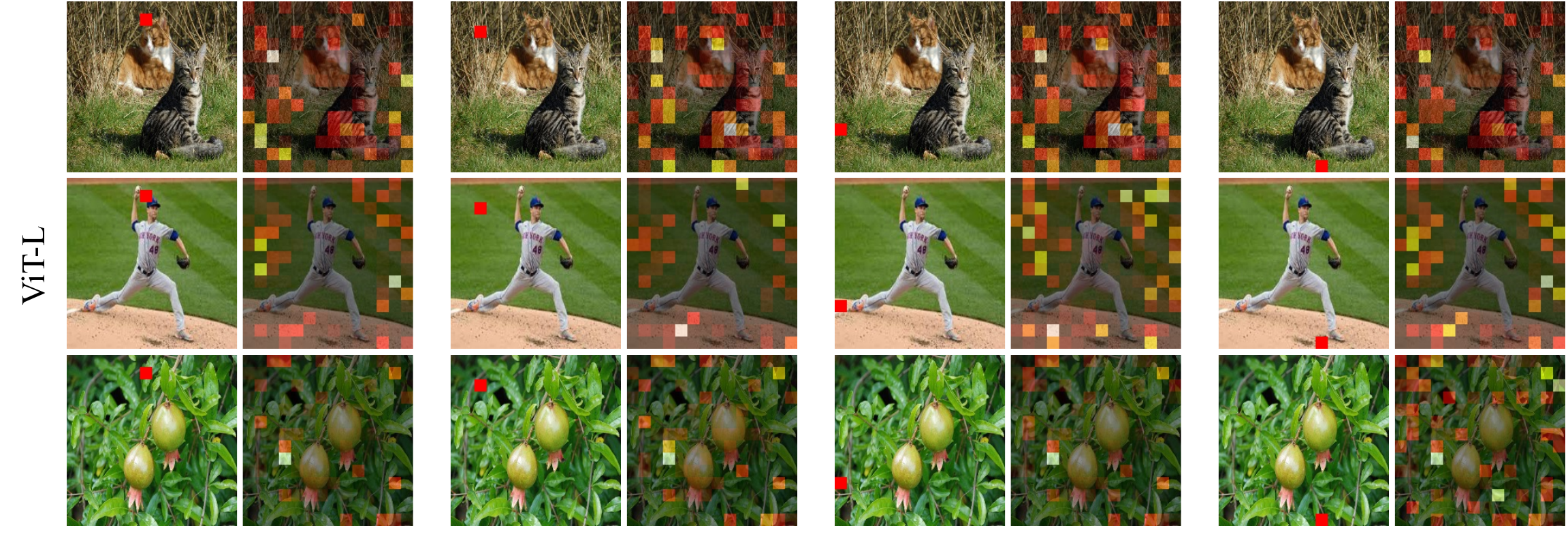}
    \includegraphics[width=0.88\linewidth]{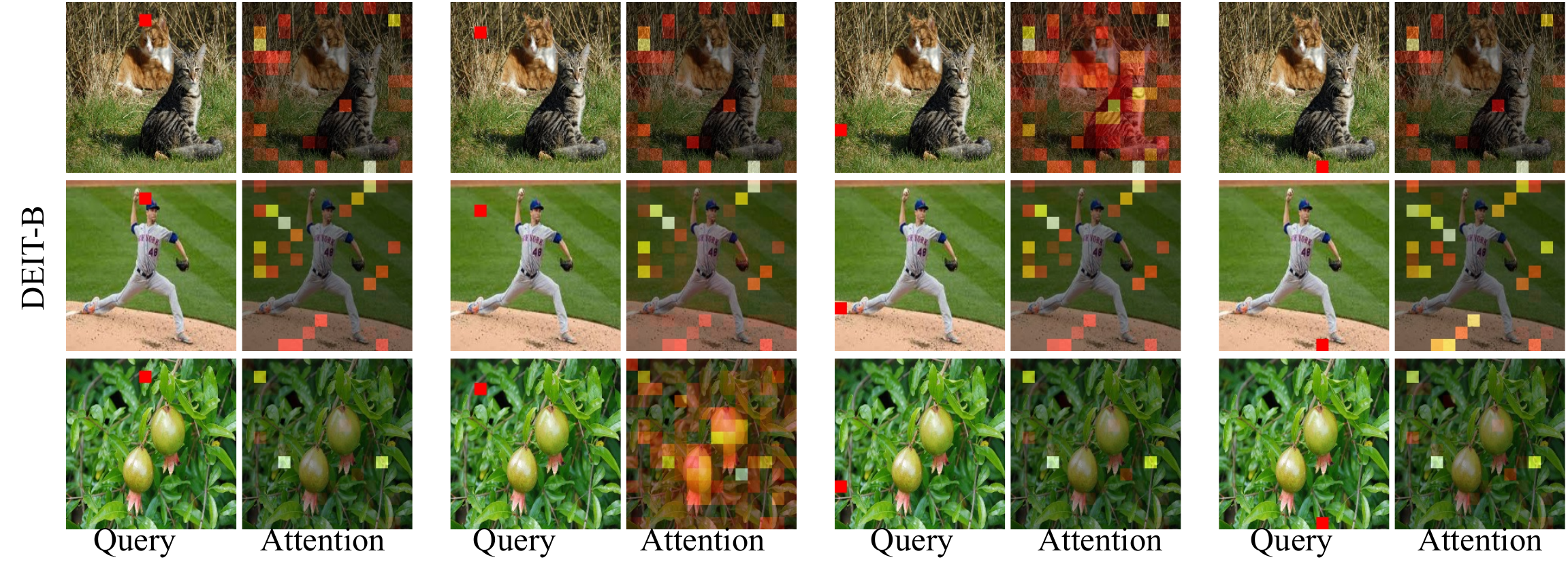}
    \vspace{-3mm}
    \caption{More attention maps of ViT variants. ViTs show a query-irrelevant behavior in general, while some bias may happen.}
    \label{fig:supp1}
    \vspace{-1mm}
\end{figure*}

\paragraph{Replacing enhanced global context with conv or attention.}  Next, we replace the global context module with multi-head attention or convolution. It is relatively hard to compare FCViT and the related replacements fairly since the parameters and flops may change a lot. To this end, we tailor each variant to achieve a roughly fair comparison. 
For the self-attention variant, we use adaptive-average-pooling (to a resolution of 7) to reduce the complexity. We consider 4 heads and 32 dimensions for each head. For convolution variants, we repeat convolution twice in each block and increase the block number in stage 3 to match the computations. 

The results are presented in Table~\ref{tab:replace}. Using even fewer parameters and FLOPs, our FCVIT-Tiny still achieves better results than the self-attention and convolution variants. Notice that convolution variants are deeper, and the self-attention variant introduces avg-pooling (which can be considered as convolution). All these variants train slower than our FCViT (around 1.2x to 1.3x time cost). Results showcased that our FCViT still outperforms all related variants, demonstrating the effectiveness of our method.

\section{More Visualizations} 

\paragraph{Similarity map in FCViT.} We also present more token-global similarity results on different examples, as shown in Fig~\ref{fig:supp_sim}. We use the pre-trained FCViT-B24 for illustration.

\paragraph{Attention Map in ViTs.} To further support our motivation, we plot more attention maps in ViTs on more examples. The results are shown in Fig.~\ref{fig:supp1}.

\end{document}